\documentclass[article]{article}

\usepackage{PRIMEarxiv}

\usepackage[utf8]{inputenc} 
\usepackage[T1]{fontenc}    
\usepackage{hyperref}       
\usepackage{url}            
\usepackage{booktabs}       
\usepackage{amsfonts}       
\usepackage{nicefrac}       
\usepackage{microtype}      
\usepackage{lipsum}
\usepackage{fancyhdr}       
\usepackage{graphicx}       
\graphicspath{{media/}}     
\usepackage{float}
\usepackage{here}

\title{Suvach - Generated Hindi QA benchmark
}

\author{
  Vaishak Narayanan \\
  \texttt{vaishakn.stats}@gmail.com  \\
  \And
  Prabin Raj KP \\
  \texttt{prabinraj.kp18@gmail.com}  \\
  \And
  Saifudheen Nouphal \\
  \texttt{saifudheennouphal@gmail.com}
}

\begin{document}
\maketitle

\begin{abstract}
Current evaluation benchmarks for question answering (QA) in Indic languages often rely on machine translation of existing English datasets. This approach suffers from bias and inaccuracies inherent in machine translation, leading to datasets that may not reflect the true capabilities of EQA models for Indic languages. This paper proposes a new benchmark specifically designed for evaluating Hindi EQA models and discusses the methodology to do the same for any task. This method leverages large language models (LLMs) to generate a high-quality dataset in an extractive setting, ensuring its relevance for the target language. We believe this new resource will foster advancements in Hindi NLP research by providing a more accurate and reliable evaluation tool.
\end{abstract}

\section{Introduction}

Recent breakthroughs in Large Language Models (LLMs), particularly those following the advent of ChatGPT, were transformative. LLMs and Artificial Intelligence (AI) as a whole have the capability to revolutionize sectors like education, healthcare, and governance, especially in densely populated nations like India. However, advancements in LLMs are often skewed towards the English language. This is particularly true for smaller, more accessible models, where the training data does not contain many tokens from Indic languages. This disparity hinders AI progress for speakers of under-resourced Indic languages, a population group encompassing over one-sixth of the world's inhabitants. Consequently, further fine-tuning or training entirely new models from scratch becomes necessary in these languages. Recent fine-tuned models like OpenHathi ,Airavata \cite{gala2024airavata}, and Tamil-LLAMA \cite{balachandran2023tamil} are a step in this direction.

But evaluating their effectiveness in low-resource Indian languages remains a challenge. With the emergence of Indic LLMs, the need for a dedicated benchmark tailored to evaluate these models becomes increasingly important. However, a significant scarcity of Indic language data persists, hindering benchmark development and training of AI models in general. This data gap can be addressed through either machine translation (MT) or generation techniques. While machine translating existing benchmarks is prevalent, it is known to amplify biases and other quality issues (Vanmassenhove et al., 2021) \cite{vanmassenhove2021machine} and most MT models are sentence-level which leads to losing contextual information. Still, generating a benchmark for Indic languages instead of machine translation of English based benchmarks remained a challenge as it was not cost-effective, and comparatively higher translation quality offered by IndicTrans2 (Gala et al., 2023)\cite{gala2023indictrans2}. Moreover, ChatGPT’s\cite{OpenAI2022} generation quality in Hindi was not up to the mark (Ahuja et al., 2023)\cite{ahuja2023mega}.

While machine translation offers a temporary solution, it is not a sustainable approach for developing long-term, large-scale benchmarks across all Indian languages. This highlights the critical need for generating benchmarks specifically designed for these languages, potentially leveraging the capabilities of LLMs themselves. Our manual inspection revealed that LLM outputs improve in quality when accompanied by contextual prompts. Recent models like Gemini 1.5 \cite{reid2024gemini} have shown human level understanding of an extremely low resource language called Kalamang using a grammar manual for the language. This finding suggests a promising avenue for LLM-powered benchmark creation for low-resource languages.

This paper introduces Suvach, a novel benchmark for extractive question answering (QA) tasks in Hindi. Suvach capitalizes on the capabilities of cutting-edge LLMs, to generate a comprehensive dataset specifically tailored to the needs of the Hindi language. By circumventing the potential pitfalls of machine-translated data, Suvach establishes a more robust evaluation environment for Indic LLMs. Furthermore, the methodology outlined here can be generalized for benchmark creation across various tasks. In this instance, we have chosen to focus on extractive question answering for multiple-choice questions (MCQs).

\section{Related Works}
\label{sec:headings}

English benchmarks such as MMLU (Hendrycks et al., 2021), Hellaswag (Zellers et al., 2019)\cite{zellers2019hellaswag}, ARC (Clark et al., 2018)\cite{clark2018think}, Winogrande (Sakaguchi et al., 2020) \cite{sakaguchi2021winogrande} and BoolQ (Clark et al., 2019) \cite{clark2019boolq} were translated using the IndicTrans2 model (Gala et al., 2023) \cite{gala2023indictrans2} and used for the evaluation of some models. These benchmarks contain evaluation on various qualitative (e.g., law, philosophy, and history) and quantitative topics (e.g., physics, computer science, and mathematics), as well as knowledge about human behavior and society (e.g., economics, sociology, politics, geography, and psychology).

Alongside IndicTrans2 model, Gala et al. (2023)\cite{gala2023indictrans2} introduced IN22, a comprehensive benchmark for evaluating machine translation (MT) performance across all 22 Indian languages. IN22 offers two unique subsets: IN22-Gen, focusing on high-quality translations from diverse domains in Indian contexts, and IN22-Conv, containing translations of everyday conversational sentences for improved MT evaluation in realistic scenarios.

An important future direction would involve creating equivalent benchmarks in the native language instead of solely relying on translations. Hence, using larger LLMs for question generation is crucial.

\section{Hindi – Extractive QA Benchmark}

\subsection{Workflow description}
Workflow used for generation (See Figure \ref{fig:flow}) can be broken down into following steps:

\begin{figure}[t] 
  \centering 
  \includegraphics[width=1\textwidth]{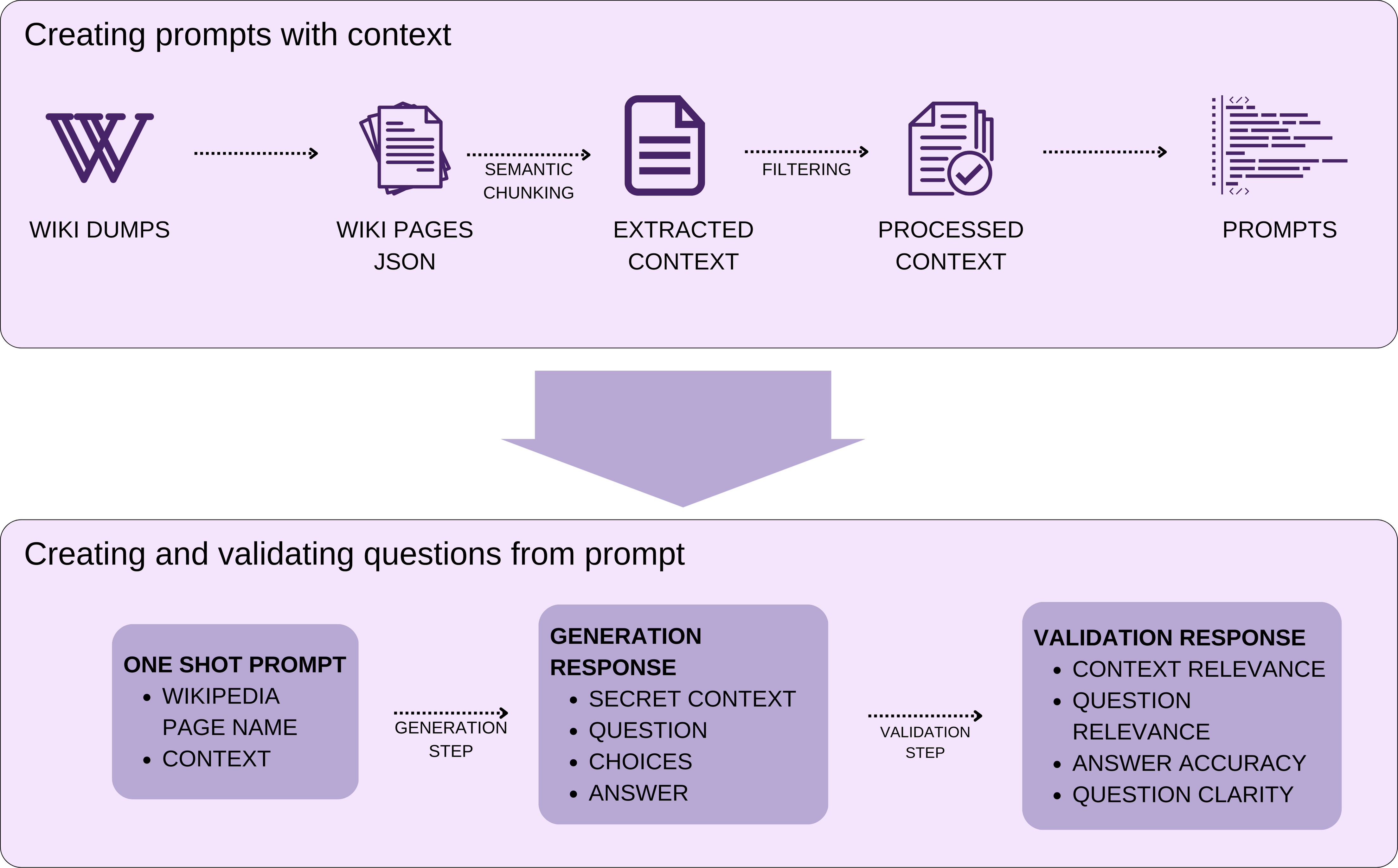} 
  \caption{End to End workflow used for question generation and validation} \label{fig:flow} 
\end{figure}

\subsubsection{Creating prompts with context}
In this step, wiki dumps are used to create prompts that include context for the question to be answered. The wiki dumps for Hindi are preprocessed into json files with page title and page content. 

The extracted page contents will be chunked for creating multiple question from the same page. We will be using \cite{fullstackretrieval2024} to creates a chunk with similar text. The chunks that do not meet a specific criteria like length, are filtered. Finally, One-Shot Prompts are created which includes the chunks as context. The format of prompt prepared is given in the appendix (Figure ~\ref{appendix:qgen}).

\subsubsection{Creating a dataset from prompts using LLM}
In this step, prompts are used by the LLM to generate a response. Then, the generated question and answer pairs are validated to remove cases where the generated response might be wrong. Validation is done using the LLM itself with following questions:

\begin{enumerate}
  \item \textbf{Context Relevance:} Does the context contain enough information to answer the question?
  \item \textbf{Question Relevance:} Is the question related to the context?
  \item \textbf{Answer Accuracy:} Is the correct answer accurately marked based on the context provided?
  \item \textbf{Question Clarity:} The question should not lead to multiple interpretations. Is the question clear and unambiguous?
\end{enumerate}

The format of prompt prepared is given in the appendix (Figure ~\ref{appendix:qval}).

\subsection{Dataset description}
This dataset consists of over 100k question answers in Hindi, with 1200 tokens per question on average. The data used for generating the question is the title of the Wikipedia page that the question was scrapped from and a chunk of text created from the given Wikipedia page (i.e., the context). The generated part of data contain Secret Context, Question, Choices, Answer, and Description.

\begin{figure}[t] 
  \centering 
  \includegraphics[width=1\textwidth]{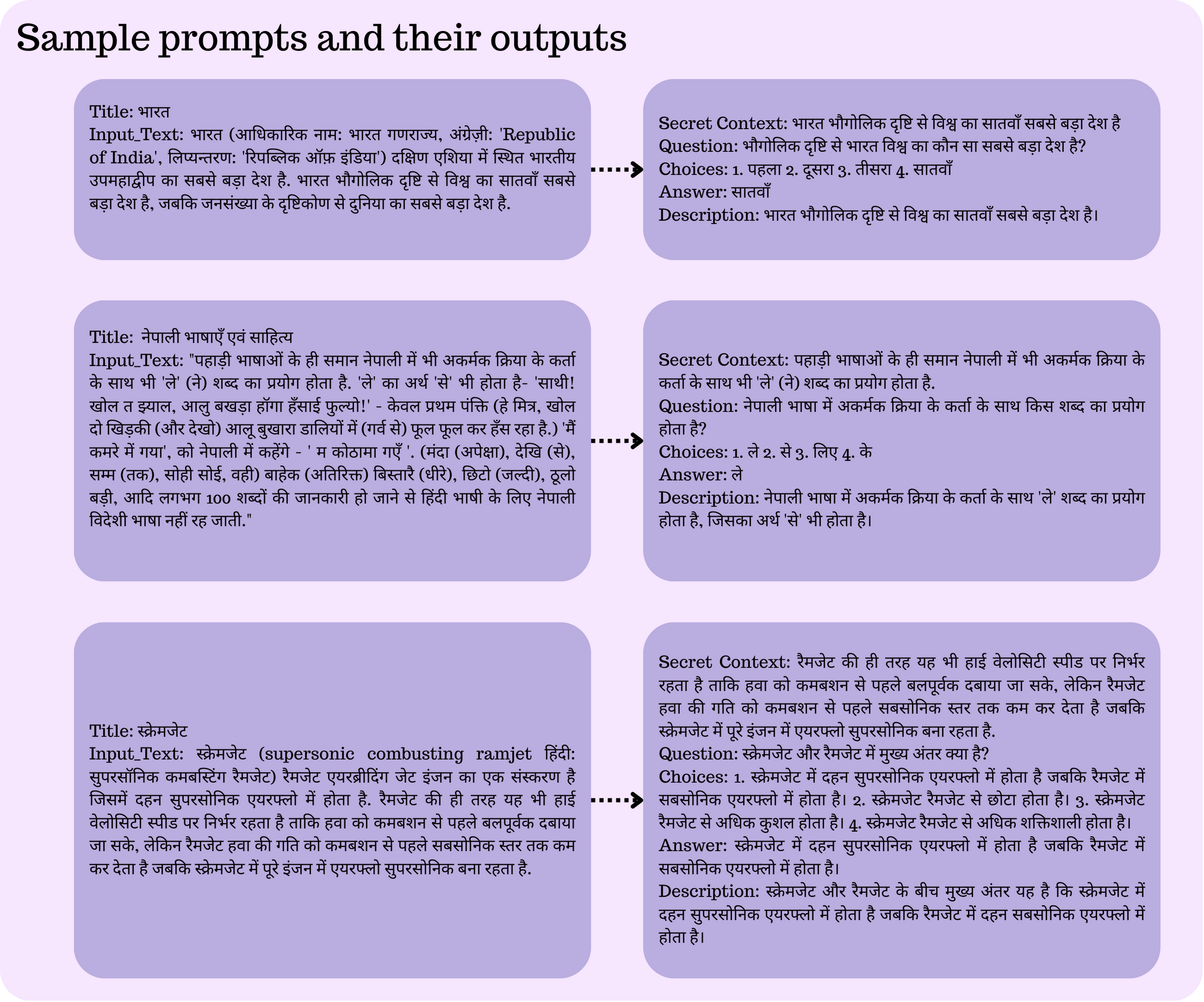} 
  \caption{Samples of chunks and the generated questions} \label{fig:qsample} 
\end{figure}

 The question will be accompanied with 4 Choices and one and only one of them would be the correct answer. For improving generation quality, a retrieval step is added to extract a chunk of text relevant to the question before generating the tokens of question itself. For improving consistency, a description is also asked to be in the response.

\begin{enumerate}
  \item \textbf{Question only :} Use only the question in evaluation prompt.
  \item \textbf{Question with context:} Provide the context along with the question in evaluation prompt.
  \item \textbf{Question with context and choices:} This would be the most easy task. The context and four possible answers are given in the evaluation prompt along with the question. Choose the most appropriate response.
\end{enumerate}

You can find some sample questions in Figure~\ref{fig:qsample}.

\newpage
\section{Conclusion}
A critical challenge in the field of natural language processing (NLP) is the dearth of high-quality benchmarks for low-resource languages. Traditionally, these benchmarks were constructed by machine translating existing English benchmarks. This approach suffers from two key limitations: (1) bias towards frequently used English words, and (2) loss of the richness and diversity inherent in the target language during translation. Consequently, LLM outputs are inadvertently evaluated using flawed benchmarks, hindering accurate assessment.  However, a promising solution emerges with the growing availability of large, freely accessible LLMs trained on massive datasets. These models, coupled with their extensive context capabilities, demonstrate significant potential for benchmark generation. As demonstrated using added context to maintain quality in our approach, further research is warranted to explore LLM-powered benchmark creation for various tasks and across a broader range of Indic languages.
\clearpage

\section*{Acknowledgments}
This dataset was generated using Gemini 1.0 Pro model.

\bibliographystyle{unsrt}
\bibliography{references} 

\newpage    
\section*{Appendix}
\appendix{}
\begin{figure}[h] 
  \centering 
  \includegraphics[height=1\textwidth]{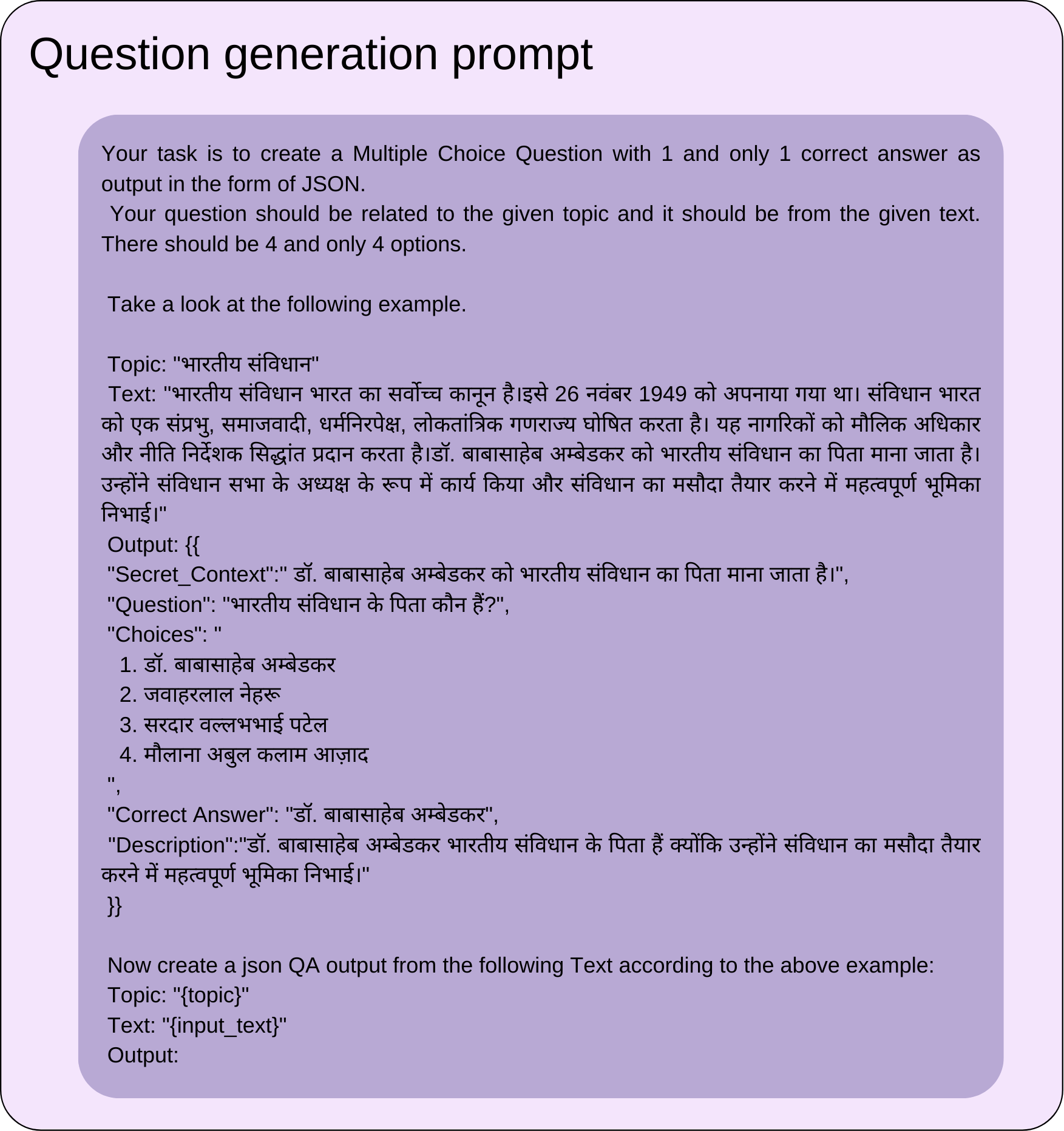} 
  \caption{The prompt used for generation of question} \label{appendix:qgen} 
\end{figure}

\begin{figure}[H] 
  \centering 
  \includegraphics[width=0.8\textwidth]{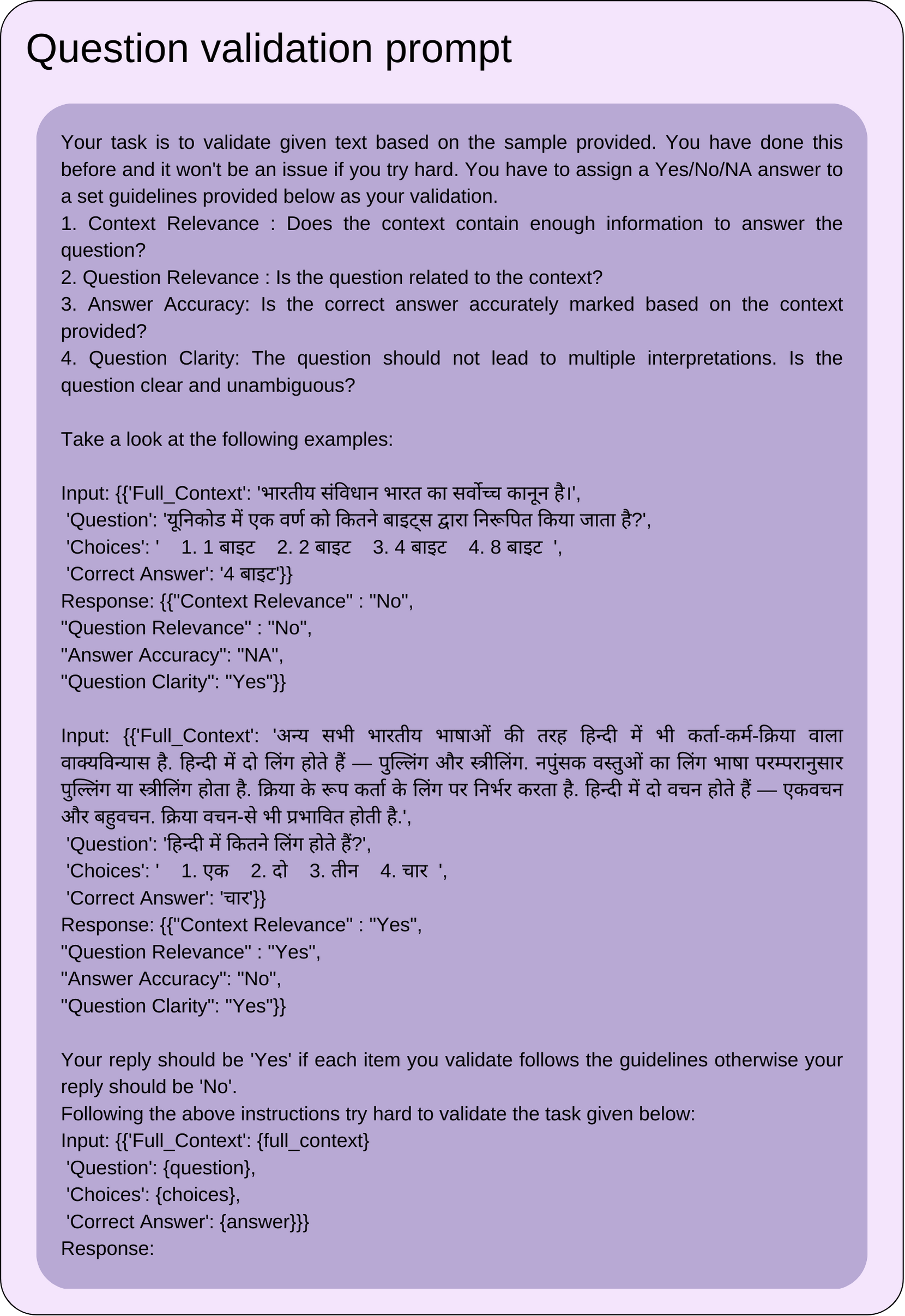} 
  \caption{The prompt used for validation of the generated question} \label{appendix:qval} 
\end{figure}

\end{document}